\documentclass[a4paper,twoside]{article}

\usepackage{epsfig}
\usepackage{subcaption}
\usepackage{calc}
\usepackage{amssymb}
\usepackage{amstext}
\usepackage{amsmath}
\usepackage{amsthm}
\usepackage{multicol}
\usepackage{pslatex}
\usepackage{apalike}
\usepackage{color}
\usepackage{adjustbox}
\usepackage{hyperref}
\usepackage{multirow}
\usepackage{SCITEPRESS}     

\def\eg{\emph{e.g.}}

\def\etal{\emph{et al.}}
\def\bev{\emph{bird's-eye view}}
\def\birdslam{\emph{BirdSLAM}}

\begin{document}

\title{BirdSLAM: Monocular Multibody SLAM in Bird's-Eye View}

\author{\authorname{Swapnil Daga\sup{1}, Gokul B. Nair\sup{1}, Anirudha Ramesh\sup{1}, Rahul Sajnani\sup{1}, Junaid Ahmed Ansari\sup{2} and K. Madhava Krishna\sup{1}}
\affiliation{\sup{1}Robotics Research Center, KCIS, IIIT Hyderabad, India}
\affiliation{\sup{2}Embedded Systems and Robotics, TCS Innovation Labs, Kolkata, India}
}

 
\keywords{Monocular SLAM, Multibody Simultaneous Localization and Mapping, Bird's-Eye View}

\abstract{In this paper, we present \birdslam{}, a novel simultaneous localization and mapping (SLAM) system for the challenging scenario of autonomous driving platforms equipped with only a monocular camera. \birdslam{} tackles challenges faced by other monocular SLAM systems (such as scale ambiguity in monocular reconstruction, dynamic object localization, and uncertainty in feature representation) by using an orthographic (\emph{bird’s-eye}) view as the configuration space in which localization and mapping are performed. By assuming only the height of the ego-camera above the ground, \birdslam{} leverages single-view metrology cues to accurately localize the ego-vehicle and all other traffic participants in \bev{}. We demonstrate that our system outperforms prior work that uses strictly greater information, and highlight the relevance of each design decision via an ablation analysis.}

\onecolumn \maketitle \normalsize \setcounter{footnote}{0} \vfill

\section{INTRODUCTION}
\label{sec:intro}

The \emph{race to level 5 autonomy} is a thrust factor in developing accurate perception modules for driverless vehicles. A majority of such industrially-led solutions rely on a suite of sensors 
such as Lidar, GPS, IMUs, radars, cameras or different permutations of such sensors. In this paper, we deviate from this paradigm and pose a challenging research question: ``\emph{How accurately can we estimate the ego motion of a driving platform and the state of the world around it, by using only a single (monocular) camera}"? In robotics parlance, this task of estimating the ego-motion of a ``robot" and the state of its environment is referred to as simultaneous localization and mapping (SLAM)~\cite{slam1,slam2}. A generalization of the SLAM problem---known as \emph{multibody} SLAM---is of interest to us. While a conventional SLAM system only estimates the robot's ego-motion and the static scene map by using the stationary features, multibody SLAM additionally estimates every other actor's motion in the scene - hence a generalized system. 
This is of paramount importance to autonomous driving platforms, as a precise estimation of the states of other actors immensely boosts the performance of downstream tasks, such as collision avoidance and over-taking maneuver.

\begin{figure}[htbp]
    \centering
    \includegraphics[width=0.9\columnwidth]{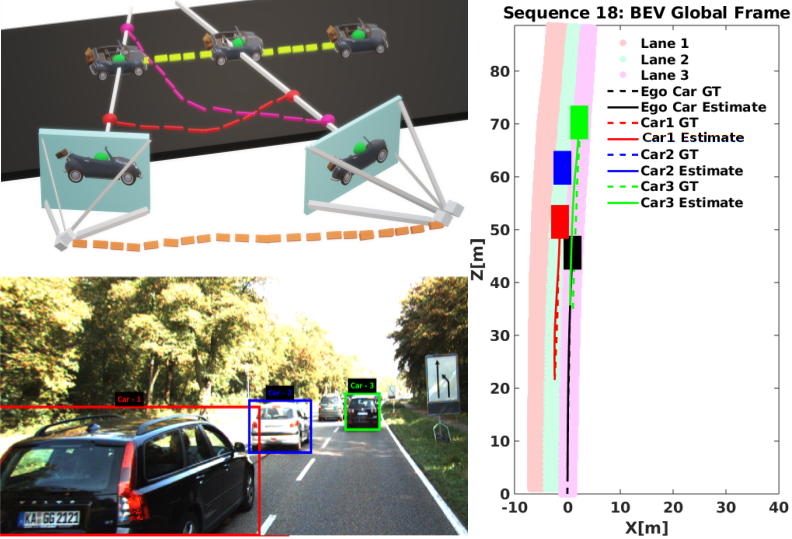}
    \caption{Top-Left: Illustration of ill-posedness of Multibody SLAM. Triangulating a moving object with a moving camera is impossible as the object has moved away by the time the second image is captured. Back projected rays intersect at highly erroneous locations. While the car moves along the yellow line, many possible trajectories (red, magenta) project to the same locations in the image. Bottom-Left and Right: BirdSLAM operates in orthographic view overcoming many nuisance factors and making optimizations simpler, thus convenient to be plugged into downstream planners.}
    \label{fig:teaser}
\end{figure}

In general, multibody SLAM is \emph{ill-posed} (i.e., does not admit a unique solution family) in moving monocular camera setup (see Fig.~\ref{fig:teaser}). This is because monocular reconstruction~\cite{orbslam2,lsd,ptam,monoslam} inherently suffers from \emph{scale factor ambiguity}. This makes it near-impossible to recover object trajectories in metric units that can be directly employed in the downstream tasks mentioned earlier. Thus, monocular cameras have so far found far fewer applications in autonomous driving stacks. In this work, we move monocular SLAM systems one step closer to downstream 
modules.

Conventional monocular SLAM systems~\cite{orb,orbslam2} detect and track sparse geometric features across input images and produce a point cloud reconstruction of the scene. These systems are faced with a plethora of issues when deployed in scenes with highly dynamic actors (\eg, traffic): consistent geometric feature matches are hard to obtain across vehicles; the passage of vehicles suddenly obstructs static scene regions with stable features; the (already ambiguous) scale of reconstruction drifts unexpectedly and rapidly. Existing approaches 
tackle some of these issues by assuming auxiliary inputs such as optical flow~\cite{vibhav2016} or depth from stereo cameras~\cite{dinesh2016,li2018stereo}. Others~\cite{multibody-kanade,two-view-multibody,multiple-motion-uncalib} pose the problem as that of \emph{factorizing} multiple motions from a 3D trajectory ``soup". Recent approaches that operate on monocular cameras are unsuitable for real-time applications~\cite{iv,cubeslam}.

In this paper, we propose \birdslam{}: a monocular multibody SLAM system tailored for typical urban driving scenarios. It operates on an orthographic view (the \bev{}), where the impact of the aforementioned ``nuisance factors" is low also making the optimizations simpler as there are less parameters to operate upon. Further, estimates in orthographic views can be directly plugged into downstream planners: a desirable quality (Fig.~\ref{fig:teaser}). By assuming that all relevant ``actions" happen on or close to the ground-plane, and leveraging single-view metrology cues, \birdslam{} enables scale-unambiguous motion estimation of the ego vehicle and other traffic participants.


\birdslam{} leverages static features available from an off the shelf SLAM system~\cite{orbslam2}, dynamic features provided by modern object detectors~\cite{mono3d,oft,pseudolidar}, and single-view metrology cues~\cite{MobileEye,songcvpr15} to formulate a scale-aware pose-graph optimization problem in \bev{}. This can be solved using off-the-shelf pose-graph optimization toolboxes~\cite{g2o,ceres-solver,gtsam}. We demonstrate that \birdslam{} outperforms existing full 6-DoF SLAM frameworks and provide an ablation analysis to justify our design choices.




In summary, \birdslam{} accurately estimates ego-motion and other vehicle trajectories in \bev{} over \emph{long sequences} in \emph{real-time}, mitigating the various nuances of traditional 6-DoF SLAM frameworks for dynamic scenes. Additional qualitative results can be found in our video. We observe that a 3-DoF SLAM results on an SE(2) representation of real road plane scenarios compare well with the traditional 6-DoF SLAM results. This simplifies the optimization parameterization thus contributing positively to reduced runtime as shown in Sec.\ref{subsubsec:runtime}. while not sacrificing on the Absolute Translation Error(ATE).


\section{RELATED WORK}
\label{sec:related_works}
\paragraph*{Traditional Approaches:}
The traditional approaches to solving the SLAM problem's multibody counterpart are based on separating multiple motions~\cite{multibody-kanade,multibody-sam,two-view-multibody,multiple-motion-uncalib,multi-body-seg} from a given set of triangulated points. Other traditional approaches included solving for relative scale for each vehicle in the scene~\cite{two-view-outlier,kundu2011,Namdev}. The relative scale reconstruction in most of such approaches is not in metric scale.

\paragraph*{Deep Learning based Approaches:} Deep learning based methods such as Reddy \etal~\cite{dinesh2016} and Li \etal~\cite{li2018stereo} leverage the improvement in object detection in deep learning approaches over traditional approaches to improve the multibody SLAM. However, these two methods use stereo cameras, thus not facing the problem of scale ambiguity, which is prevalent in monocular settings.

\paragraph*{Recent Approaches:}
A more recent approach to the multibody SLAM problem in a monocular setting is proposed by Nair \etal~\cite{iv} which relies on batch-based pose-graph optimization in 6 DoF. The optimization framework used in it cannot be applied in a real-time setting. Another recent framework Cubeslam~\cite{cubeslam} uses object representations in 6 DoF to improve ego vehicle trajectories; however, the problem is not cast into a dynamic setting, and dynamic participant's trajectories are not shown explicitly.
\begin{figure*}[!t]
\centering
\includegraphics[width=\linewidth]{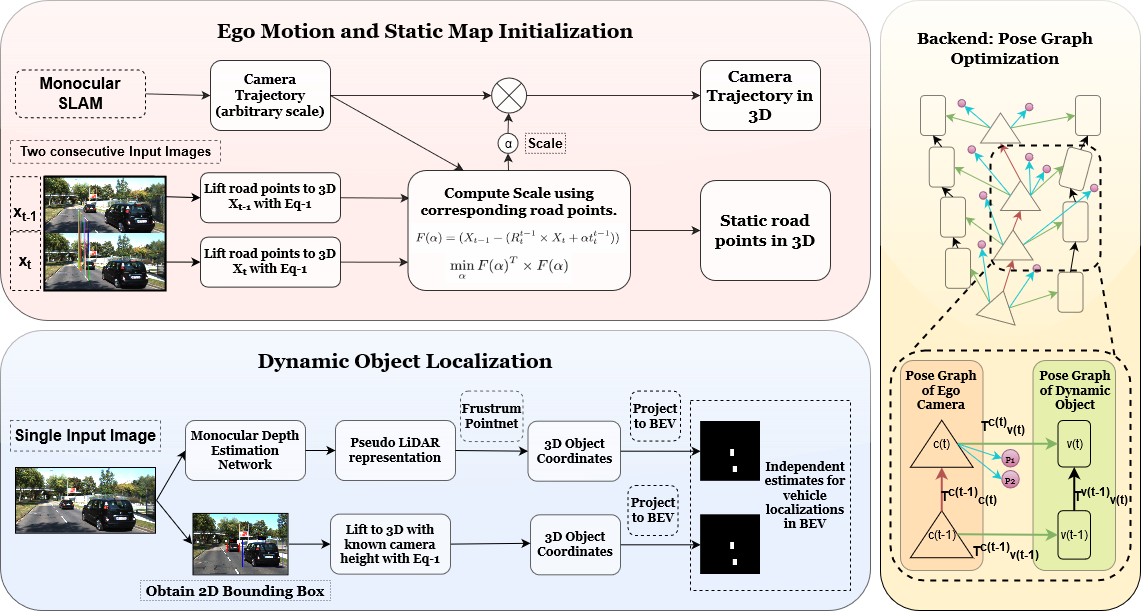}
\caption{Pipeline: The Ego Motion and Static Map Initialization block illustrates the generation of static road points in 3D, in addition to how we obtain camera trajectory in metric scale. The Dynamic Object Localization block illustrates our approach to obtain two independent sources of localization of other dynamic objects in frame in Birds-Eye View (BEV). Detailed explanation, and mathematical representation of these blocks can be found in Sec.~\ref{subsec:vehloc}. The Backend Pose Graph Optimization block uses the results obtained from Ego Motion and Static Map Initialization block, and the Dynamic Object Localization block to create a pose graph as illustrated. In the pose graph formulation, the red, black, green, and blue arrows show the CC, VV, CV and CP constraints as described in Sec.~\ref{subsubsec:constr}. }
\label{fig:pipeline}
\end{figure*}

\section{BIRDSLAM: MULTIBODY SLAM IN BIRD's-EYE VIEW}
\label{sec:approach}

\subsection*{Problem Formulation} Given a sequence of monocular images ${\mathcal{I}_i}, i \in {1 \cdots N},$ captured from an urban driving platform, with the camera at height H above the ground, the task of \birdslam{} is to estimate:
\begin{enumerate}
\item The ego motion of the vehicle $X_i = (x_i, z_i, \theta_i)$ at each time step, on the ground plane (assumed to be the $XZ$ plane)
\item An estimate of the motion of all other traffic participants $X^j_i = (x^j_i, z^j_i, \theta^j_i), j \in {1 .. M_i}$, where $M_i$ is the number of traffic participants detected in image $I_i$.
\item A \emph{map} $\mathcal{M}$ of the environment comprising static features on the road plane (such as lane markings etc.).
\end{enumerate}

The overall pipeline of \birdslam{} can be seen in Fig.~\ref{fig:pipeline}, where the input images are first passed through an ego motion estimation pipeline (such as an off-the-shelf SLAM system). The resulting estimates are scale-compensated by using single-view metrology cues. In parallel, traffic participants and static scene points on the ground plane are mapped to \bev{} by a pseudolidar representation~\cite{pseudolidar}. This constitutes the \emph{frontend} of \birdslam{}.

The \emph{backend} of \birdslam{} comprises a novel multibody pose-graph formulation that employs constraints of several types (CC: camera-camera, CV: camera-vehicle, CP: camera-static map point, VV: vehicle-vehicle) and optimises the pose-graph in real-time to obtain globally-consistent, scale-unambiguous multibody SLAM estimates. In the following subsections, we explain each of these components in detail.


\subsection{\birdslam{}: Frontend}

\subsubsection{Static Map Initialization}
\label{subsec:statpoint}

Accurately localizing static features in a scene is critical to the success of a feature-based monocular SLAM system. We use ORB features to obtain reliable candidate ``stable" features, and prune all those features that do not lie on the road (The ``road" region is found by running a lightweight semantic segmentation network~\cite{inplace_abn} over the input image).
Using the known camera height $H$, a road point $x_{p}^{c}$ in image space can be back-projected into the camera coordinate frame as follows ($K \in R^{3 \times 3}$ is the camera intrinsic matrix, and $\overline{n} \in R^{3}$ is a unit normal to the ground-plane ($y=0$))\footnote{\label{footnote: flat_earth_assumption}\emph{Flat-earth assumption}: For the scope of this paper, we assume that the roads are somewhat planar, i.e., no steep/graded roads on mountains. Consequently, we take normal vector $n = [0,-1,0]$ in camera frame according to KITTI's~\cite{kitti} conventions where positive x-axis is in right direction, positive y-axis is in downward direction and positive z-axis is in forward direction.}: 
\begin{equation}
X_{p}^{c} = \frac{- HK^{-1}x_{p}^{c}}{\overline{n}^{T}K^{-1} x_{p}^{c}}
\label{eqn:mobiliFormula}
\end{equation}

\subsubsection{Scale-Unambiguous Ego-Motion Initialization}
\label{subsec:egomot}

We use ego-motion estimates from an off-the-shelf SLAM system~\cite{orbslam2} to bootstrap our system. Typically, such estimates are scale-ambiguous. However, upon performing the static map initialization as described in Eqn.~\ref{eqn:mobiliFormula}, we obtain map points in metric scale (since the camera height $H$ is known in \emph{meters}; it resolves scale-factor ambiguity). We use a moving-median filter to scale ego-motion estimates to real-world units (typically \emph{meters}).

\subsubsection{Dynamic Object Localization}
\label{subsec:vehloc}

\begin{figure}[!htbp]
    \centering
    \begin{tabular}{cc}
        
        \raisebox{0.15\height}{\includegraphics[width=0.5\linewidth]{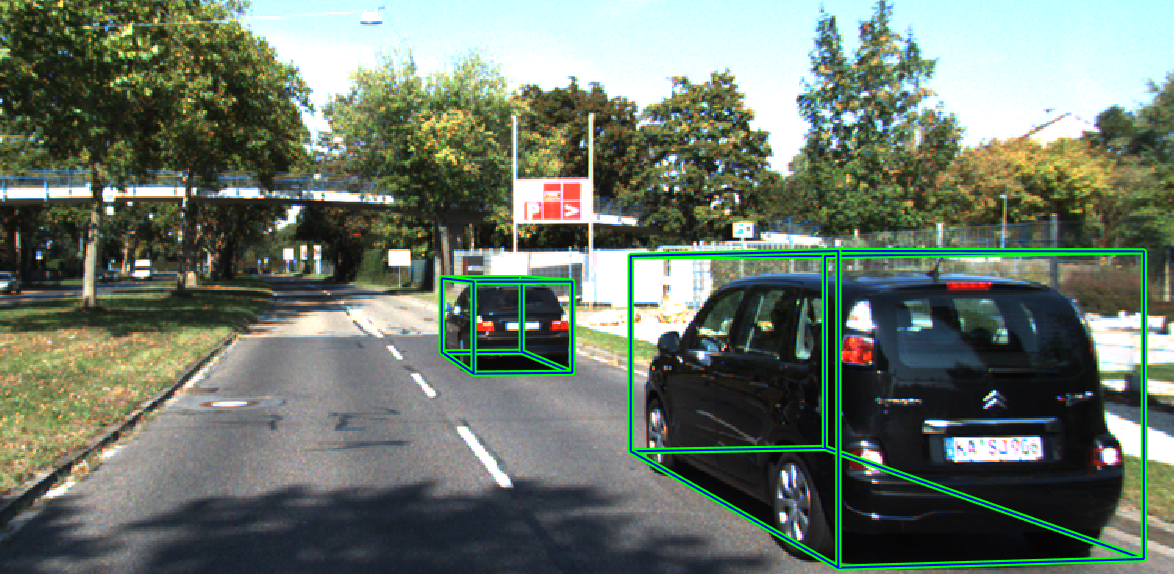}}
        & \raisebox{0.0\height}{\includegraphics[width=0.3\linewidth]{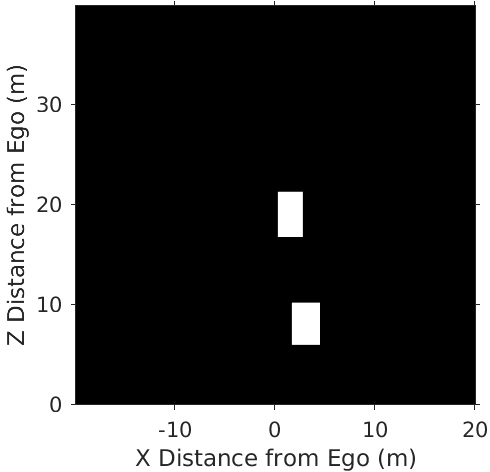}}
        
    \end{tabular}
    \caption{Vehicle Localisation in bird's-eye view: The left image shows the 3D bounding box output and the right image shows the tight bounding boxes for the cars we obtain in bird's-eye view in camera frame in metric scale from the procedure described in Sec.~\ref{subsec:vehloc}. The camera center for the right image is at (0,0) of XZ plane facing towards positive Z axis.}
    \label{fig:pseudolidar_qualitative_results}
\end{figure}

Dynamic traffic participants are the root cause of several monocular SLAM failures. In \birdslam{}, we explicitly account (and track!) other vehicles in the scene to provide state estimates that can be directly fed to a downstream planning module.
In particular, we employ a monocular depth estimation network~\cite{monodepth2} and compute a pseudolidar representation~\cite{pseudolidar} using the output depth map. The pseudolidar output is then passed to a Frustum-PointNet~\cite{frustum_pointnet} to localize vehicles in 3D (see Fig.~\ref{fig:pseudolidar_qualitative_results}). We back project these vehicles localized in 3D to \bev{} using Eqn.~\ref{eqn:mobiliFormula}. We also make use of Eqn.~\ref{eqn:mobiliFormula} on the bottom-center of 2D detection of vehicles in the camera frame as a second unique source of dynamic object localization. 

The above module is used to initialize our pose-graph defined in SE(2) in Sec.~\ref{subsubsec:bmvc}. For initializations to our pose-graph in our baselines in SE(3) in Sec.~\ref{subsubsec:baseline2}, we make use of the shape-prior based approach~\cite{murthy2017shape,murthy2017reconstructing,ansari2018earth} for vehicle localizations in the camera's coordinate system.

\subsection{Generating Amodal Lane Point Clouds in Camera Frame}
\label{subsec:lane_enet}

\begin{figure}[!htbp]
    \centering
    \begin{tabular}{cc}
        
        \multicolumn{2}{c}{\includegraphics[width=0.7\linewidth]{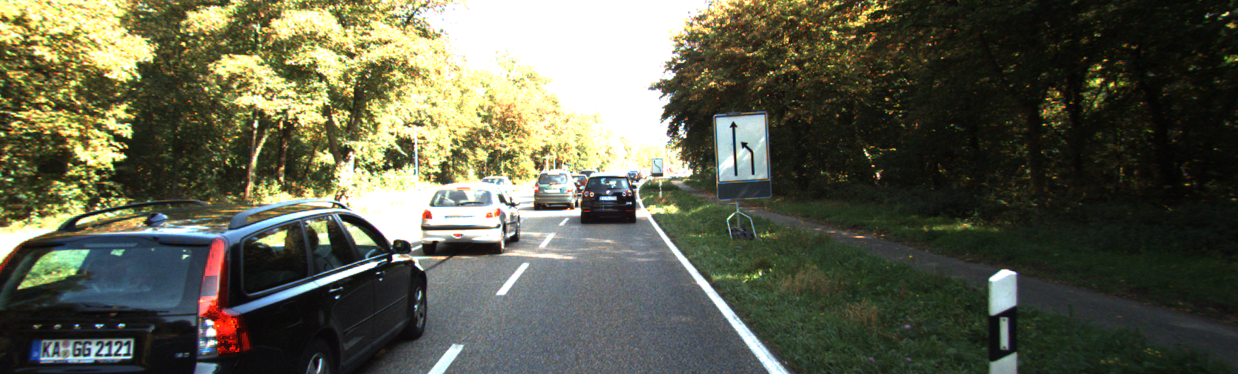}} \\
        \includegraphics[width=0.3\linewidth]{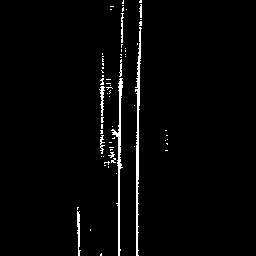} & \includegraphics[width=0.3\linewidth]{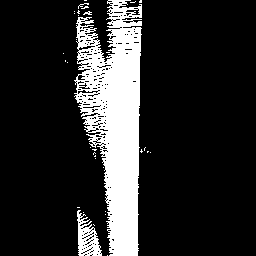} \\
        \includegraphics[width=0.3\linewidth]{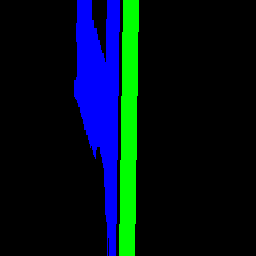} &
        \includegraphics[width=0.3\linewidth]{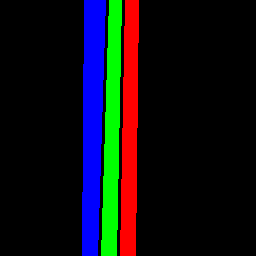}
        
    \end{tabular}
    \caption{Row 1: A frame in KITTI with 3 lanes, one of the lanes being occluded due to obstructions. Row 2, Col 1: "Lane Boundary" occupancy grid pointcloud in bird's-eye view. Row 2, Col 2: "Road" occupancy grid pointcloud in bird's-eye view. Row 3, Col 1: Lane Pointcloud output in bird's-eye view without ENet. Row 3, Col 2: Amodal Lane Pointcloud in bird's-eye view obtained after following procedure in Sec.\ref{subsec:lane_enet}.}
    \label{fig: amodal_lane_qualitative}
\end{figure}

We initialize point clouds in the camera frame using a monocular depth estimation network~\cite{monodepth2}. Using odometry over a window of $W$ frames, we aggregate sensor observations over time to generate a more dense and noise-free point cloud. To tackle noise in monocular depth estimations, we pick points up to a depth of $5m$ from the camera and then aggregate depths over a larger window size ($\approx 40-50$ frames) to compensate for its narrow field of view. This dense point cloud is then projected to an occupancy grid in the \emph{bird's eye view}. We use a state-of-the-art semantic segmentation network~\cite{inplace_abn} to segment each frame and aggregate the \emph{"road"} and \emph{"lane boundary"} prediction point clouds into \emph{separate} occupancy grids (see Fig.~\ref{fig: amodal_lane_qualitative}). To achieve more robustness, we apply additional filtering on both of the above occupancy grids by retaining only the patches with more foreground cells than a given threshold in its $m \times m$ neighborhood.

We feed these \emph{"road"} and \emph{"lane boundary"} occupancy grids into ENet~\cite{enet} to get amodal \emph{"road"} and \emph{"lane boundary"} point clouds in their respective occupancy grids. The above method is especially useful when occlusion from dynamic objects in the scene hinders input data generation. We further apply morphological post-processing techniques like \emph{opening} and \emph{closing}, followed by \emph{hough line transform}. We get segregated amodal lane point clouds in the camera frame in an occupancy grid for each monocular image as shown in Fig.~\ref{fig: amodal_lane_qualitative}. These lane point clouds are used by our SE(2) (Sec.~\ref{subsubsec:bmvc}) and SE(3) approach (Sec.~\ref{subsubsec:baseline2}) to fix lateral drifts in the optimization as explained in Sec.~\ref{subsubsec:constr}.

\subsection{\birdslam{} Backend: Pose-Graph Optimization}
\label{subsec:posegraph}

We present a lightweight \emph{online} pose-graph formulation that incorporates constraints from multiple entities in the scene (egovehicle, other vehicles, static map features). Each of these constraints contributes a \emph{cost-function} to the optimization process, which we explain below.

\subsubsection{Cost Function}
\label{subsubsec:cost}

Following g2o terminologies, the \emph{estimate} $T^W_S \in \mathbb{L}$ characterizes pose for \emph{node} $S$ in global frame $W$. Here, $\mathbb{L}$ represents the Lie Group in which the respective transformations are defined which could be $SE(2)$ or $SE(3)$. The \emph{measurement} $T^{S}_{D} \in \mathbb{L}$ denotes a \emph{binary-edge} from source node $S$ to destination node $D$ effectively constraining the respective \emph{estimates}. This can be represented mathematically as the following transform:

\begin{equation}
\Upsilon_{SD} = (T_{D}^{S})^{-1} (T_{S}^{W})^{-1} (T_{D}^{W})
\label{eqn:g2oCost}
\end{equation}

We also use \emph{unary-edges} between agent \emph{node} and stationary scene-landmarks $p$ located at $X_{p}^{W} \in \mathbb{R}^{3}$ in the global frame $W$. Here, the agent $A$ could be ego-camera or the dynamic object in scene. This does not constrain the orientation of the agent. The resultant transform between a agent node $A$ with translation vector $tr_{A}^{W} \in \mathbb{R}^{3}$ and a world landmark $p$ in global frame can be shown as:

\begin{equation}
\Psi_{A} = tr_{A}^{W} - X_{p}^{W}
\label{eqn:g2oCostUnary}
\end{equation}

Our formulation also includes a positive semi-definite inverse covariance matrix or an \emph{information matrix} in each edge's parameterization, shown as $\Omega_{E} \in \mathbb{R}^{N \times N}$ where $N \in Z$ is the number of degrees of freedom the specific edge $E$ affects. We exploit this to convey confidence of each constraint. We do so by scaling $\Omega_{E}$ upto the \emph{effective information matrix} $\overline{\Omega}_{E}$ by a factor $\lambda \in \mathbb{R}$ as: 

\begin{equation}
\overline{\Omega}_{E} = \lambda  \Omega_{E}
\label{eqn:infoScaling}
\end{equation}

From the transforms in Eqn.~\ref{eqn:g2oCost} and Eqn.~\ref{eqn:g2oCostUnary}, we obtain $e^{s} \in \mathbb{R}^{1 \times N}$ by extracting the translation vector directly, and the yaw angle(for $SE(2)$) or the axis-angle rotations(for $SE(3)$). Given the \emph{information matrix} $\Omega^{s} \in \mathbb{R}^{N \times N}$, we obtain the final cost function for either a \emph{unary} or a \emph{binary-edge}  as:

\small
\begin{equation}
\mathbf{F^{s}} = (e^{s}) (\Omega^{s}) (e^{s})^{T} 
\label{eqn:finalcost}
\end{equation}
\normalsize
 
 



\subsubsection{Constraints}
\label{subsubsec:constr}

\begin{itemize}
    \item \textbf{Exploiting dynamic cues from vehicles in the scene:} We categorize our pose-graph into three sets of relationships denoting \emph{camera motion}, \emph{vehicle motion} and \emph{camera-vehicle} constraints. Each of these is obtained as a \emph{camera-camera}, \emph{vehicle-vehicle} and \emph{camera-vehicle} edge respectively in consecutive time instants $t-1$ and $t$. We obtain the final constraint for an $m$ vehicle scenario as: 
    \begin{equation}
        \begin{split}
        \mathcal{F}_{D} = \mathbf{F}_{C(t-1),C(t)} + \sum^{m}_{j=1} \mathbf{F}^{j}_{V(t-1),V(t)} \\
        + \sum^{m}_{j=1} \mathbf{F}^{j}_{C(t-1),V(t-1)} + \sum^{m}_{j=1} \mathbf{F}^{j}_{C(t),V(t)}
        \label{eqn:dyn}
        \end{split}
    \end{equation}
    
    \item \textbf{Exploiting static cues using landmarks in the environment:} We also make use of static-cues from the environment to improve agent motion by constraining with respect to the lane. We obtain a dense point-cloud $P_{l}$ for the road plane segregated for each lane based on Sec.~\ref{subsec:lane_enet}. We define a \emph{unary-edge} between an agent $A$(Ego-camera or vehicle in scene) and each point $p$ on the lane as shown by Eqn.~\ref{eqn:g2oCostUnary}. 
    \begin{equation}
        \mathcal{F}_{S} = \sum_{p} \mathbf{F}_{A,p} \forall (p \in P_{l})
        \label{eqn:stc}
    \end{equation}
\end{itemize}

Collectively, the final cost is obtained as the sum of the above Eqn.~\ref{eqn:dyn} and Eqn.~\ref{eqn:stc}.

\begin{equation}
\mathcal{F} = \mathcal{F}_{D} + \mathcal{F}_{S}
\label{eqn:collcost}
\end{equation}


The scale of the \emph{information matrix} in Eqn.~\ref{eqn:infoScaling} is such that higher the scaling($\lambda$), more effective the corresponding cost's \emph{observation} is going to be. Thus, edges with relatively more reliable \emph{observation} are given higher weights while those that bring in higher degrees of error are weighed lower. Thus, $CC$ and $CP$ constraints have the highest weight of $10000$ while $VV$ constraints have the lowest of $1$. The weight initialization provided to $CV$ constraint ranges between $1000$ and $10$. The applied weight is gauged according to the depth of the vehicle from the camera. While pseudolidar~\cite{pseudolidar} from Sec.~\ref{subsec:vehloc} dominates at lower depths, Eqn.~\ref{eqn:mobiliFormula} to 2D vehicle detection bottom-center has an upper hand for far away objects. 
\begin{figure*}[!htbp]
    \centering
    \begin{tabular}{cc}
        
        \includegraphics[width=0.6\linewidth]{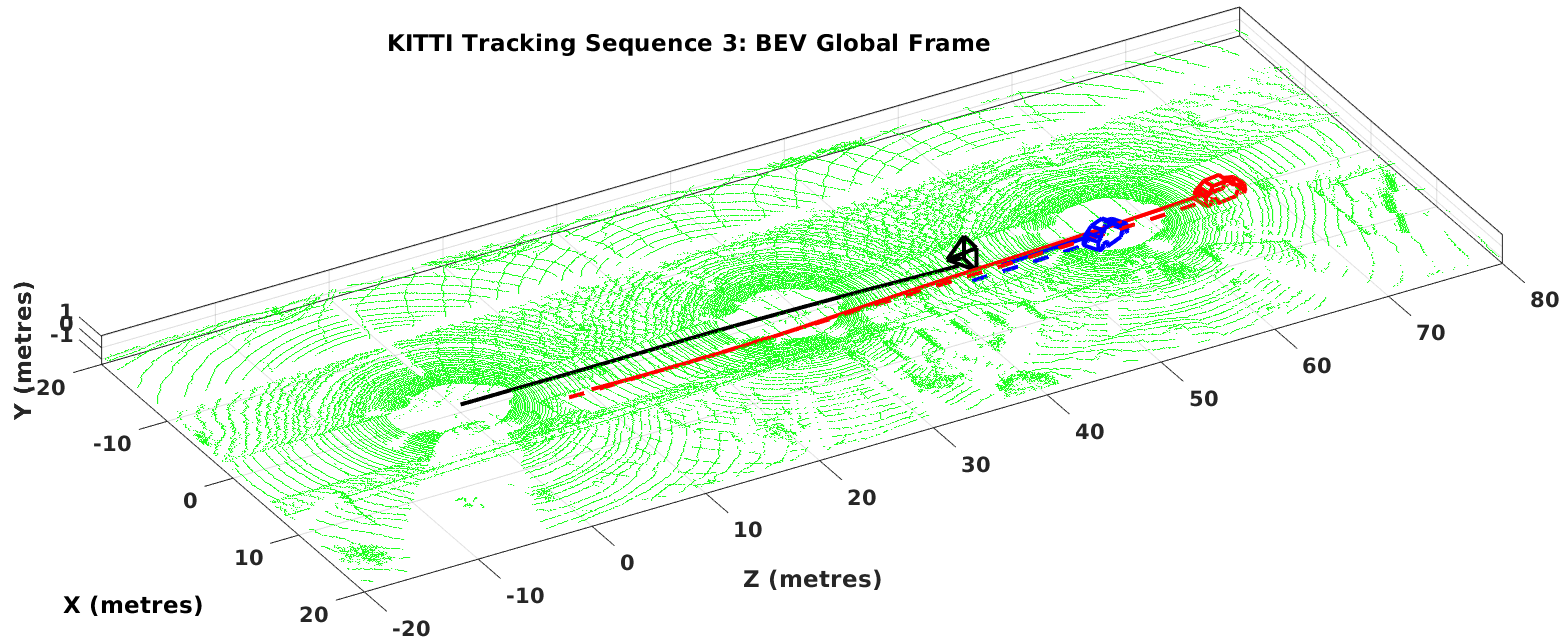} & \multirow{2}{*}[1.5in]{\includegraphics[width=0.25\linewidth]{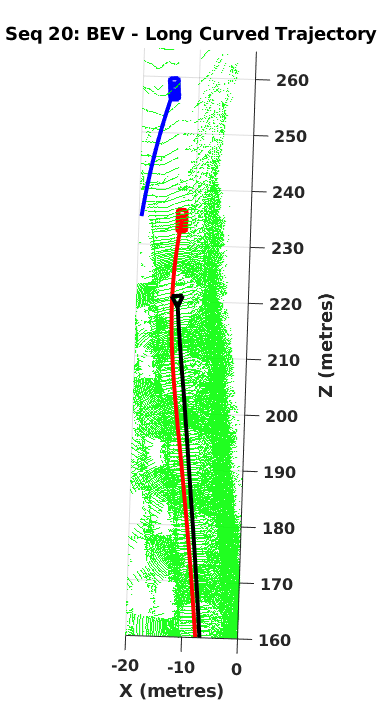}}\\
        \includegraphics[width=0.6\linewidth]{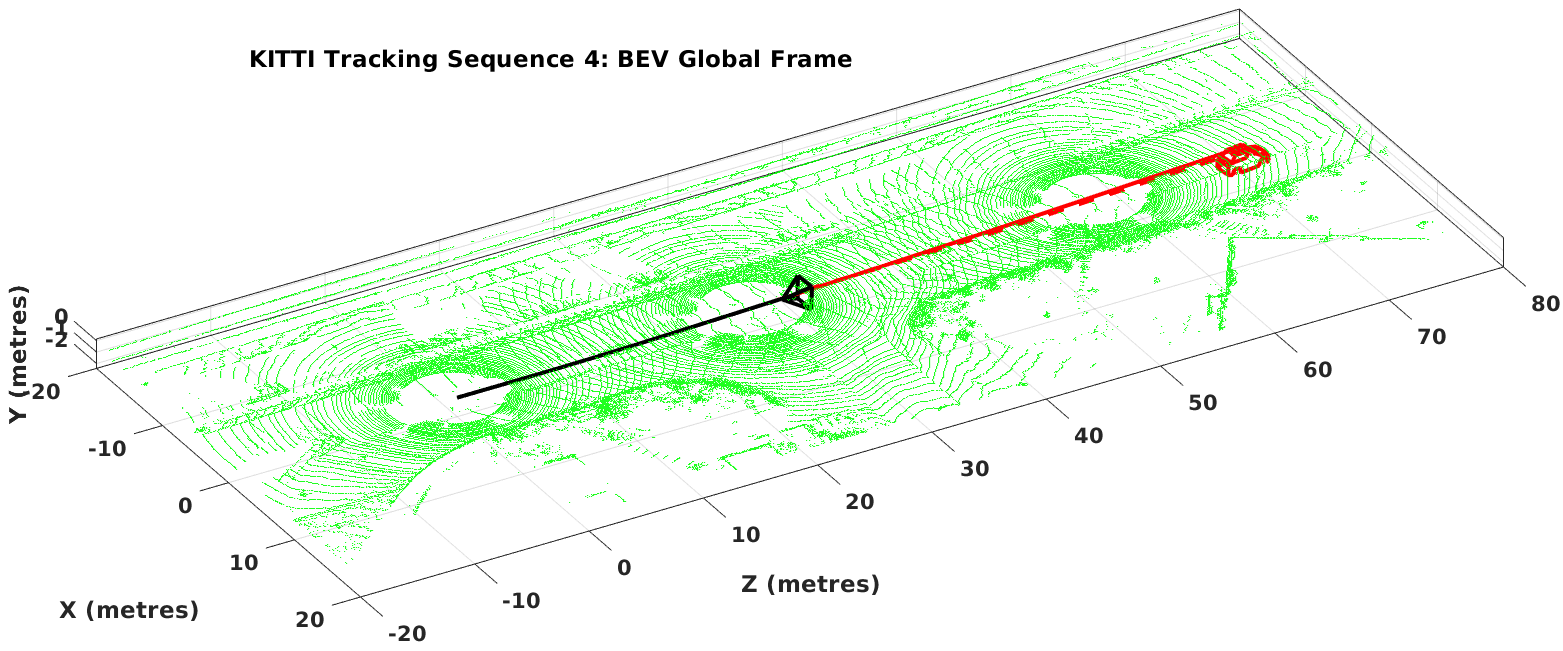} &
        
    \end{tabular}
    \caption{Qualitative Results: Visualizations of ego (black colored camera) trajectories and car object trajectories(red and blue color) for some of the KITTI sequences, shown along with surrounding lidar points in metric scale. Some of the results were obtained on very challenging sequences like curved trajectories, occluded detections etc. One such snapshot for a time instance is shown on the right for sequence 20 which had big turns in its path and some of the tracked cars were far away and occluded.}
    \label{fig:main_qualitative_results}
\end{figure*}

\section{EXPERIMENTS AND RESULTS}
\label{sec:results}

\subsubsection{Dataset} 
We perform experiments over several long KITTI-Tracking sequences~\cite{kitti}. We get ground truth localization to vehicles from the labels available with the dataset and the ground truth ego-motion from the GPS/IMU data given with the dataset.

\subsubsection{Error Evaluation}
We compute Absolute Translation Error(ATE) as the root-mean-square of error samples for each vehicle's individual frames, including the ego-vehicle in an SE(2) world. Even though the approaches evaluated in Table.~\ref{table:main_comparison_vehicle} and Table.~\ref{table:main_comparison_odometry} perform SLAM in SE(3), we project their estimated trajectories onto the ground-plane and compute their error in SE(2) setting for a fair comparison with the results of Sec.~\ref{subsubsec:bmvc}. 

\subsection{Approaches Evaluated}
\label{subsec:comparisons}

\subsubsection{Nair \etal~\cite{iv}:} 
A monocular multibody approach in SE(3) with a batch-wise pose-graph optimization formulation that resolves relationships with dynamic objects as a means of performing SLAM. 

\subsubsection{CubeSLAM~\cite{cubeslam}:} 
A monocular approach that unifies 3D object detections and and multi-view object SLAM pipelines in a way that benefits each other.

\subsubsection{Namdev \etal~\cite{Namdev}:} 
A monocular multibody VSLAM approach that obtains motion for dynamic objects and ego-camera in a unified scale. The non tractable relative scale that exists between the moving object and camera trajectories is resolved by imposing the restriction that the object motion is locally linear.

\begin{table*}[!t]
\caption{Absolute Translation Error (in meters) for localised vehicles in scene in bird's-eye view map, computed as root-mean-square error across the 2D axes.}
\label{table:main_comparison_vehicle} \centering
\begin{center}
\begin{adjustbox}{max width=\linewidth}
\begin{tabular}{|l|c|c|c|c|c|c|c|c|c|c|c|}

\hline
 & \multicolumn{11}{c|}{Absolute Translation Error (RMS) in Global Frame (meters)} \\

\hline
Seq No.& $2$ & \multicolumn{2}{c|}{$3$} & $4$ & $5$ & $10$ & \multicolumn{3}{c|}{$18$} & \multicolumn{2}{c|}{$20$} \\ \cline{1-12}

Car ID & $1$ & $0$ & $1$ & $2$ & $31$ & $0$ & $1$ & $2$ & $3$ & $12$ & $122$ \\ \hline\hline

Nair \etal~\cite{iv} & $5.01$ & $\mathbf{1.61}$	& $4.99$ & $2.14$ & $21.64$ & $3.99$ & $1.29$ & $3.45$	& $2.4$ & $9.08$ & $12.86$ \\ \hline

Namdev \etal~\cite{Namdev} & $6.35$ & $13.81$ & $11.58$ & $11.18$ & $4.09$ & $10.08$ & $3.77$ & $5.93$ & $3.72$ & $25.19$ & $23.76$ \\ \hline

\textbf{Ours} Sec.~\ref{subsubsec:baseline2} & $6.02$ & $2.20$ & $2.24$ & $\mathbf{1.77}$ & $\mathbf{1.76}$ & $3.99$ & $\mathbf{1.21}$ & $2.86$ & $\mathbf{1.23}$ & $8.96$ & $13.19$ \\ \hline \hline

CubeSLAM~\cite{cubeslam}  & $-$ & $-$ & $-$ & $-$ & $-$ & $-$ & $1.89$ & $\mathbf{2.43}$ & $7.17$ & $-$ & $-$ \\ \hline


\textbf{Ours} Sec.~\ref{subsubsec:bmvc} & $\mathbf{2.09}$ & $2.37$	& $\mathbf{2.05}$ & $2.34$ & $1.98$ & $\mathbf{3.03}$ & $1.6$ & $2.76$ & $1.6$ & $\mathbf{8.61}$ & $\mathbf{10.12}$ \\ \hline

\end{tabular}
\end{adjustbox}
\end{center}
\end{table*}

\begin{table*}[!t]
\caption{Absolute Translation Error (in meters) for ego motion in bird's-eye view map, computed as root-mean-square error across the 2D axes.}
\label{table:main_comparison_odometry} \centering
\begin{center}
\begin{adjustbox}{max width=\linewidth}
\begin{tabular}{|l|c|c|c|c|c|c|c|}

\hline
 & \multicolumn{7}{c|}{Absolute Translation Error (RMS) in Global Frame (meters)} \\

\hline
Seq No.& $2$ & $3$ & $4$ & $5$ & $10$ & $18$ & $20$ \\ \cline{1-8}

No. of Frames & $67$ & $123$ & $149$ & $101$ & $249$ & $141$ & $414$  \\ \hline\hline

Nair \etal~\cite{iv} & $2.30$ & $1.96$	& $6.49$ & $1.60$ & $10.05$ & $2.40$ & $8.85$ \\ \hline

Namdev \etal~\cite{Namdev} & $6.24$ & $11.49$ & $11.12$ & $4.08$ & $10.05$ & $3.96$ & $24.38$  \\ \hline

\textbf{Ours} Sec.~\ref{subsubsec:baseline2} & $\mathbf{2.05}$ & $1.96$ & $\mathbf{1.89}$ & $2.22$ & $3.16$ & $2.36$ & $9.05$  \\ \hline \hline

CubeSLAM~\cite{cubeslam}  & $-$ & $-$ & $-$ & $-$ & $-$ & $2.99$ & $-$ \\ \hline


\textbf{Ours} Sec.~\ref{subsubsec:bmvc} & $2.25$ & $\mathbf{1.78}$	& $6.60$ & $\mathbf{1.58}$ & $\mathbf{2.99}$ & $\mathbf{1.60}$ & $\mathbf{8.81}$ \\ \hline

\end{tabular}
\end{adjustbox}
\end{center}
\end{table*}

\subsubsection{Batch Optimized Baseline in SE(3) with Scale-ambiguous ORB Odometry:} \label{subsubsec:baseline2}
A monocular multibody approach in SE(3) similar to Nair \etal~\cite{iv} but the camera nodes are fed with scale-ambiguous ORB~\cite{orbslam2} initialization. We show that the optimizer itself is able to pull scale-ambiguous odometry to \emph{metric scale} without relying on any prior scale correction like Sec.~\ref{subsec:egomot}. We incorporate stationary landmarks into this pipeline in the form of a dense lane point cloud for each lane obtained from Sec.~\ref{subsec:lane_enet} applied in a batch version to correct for the lateral drift contributed to by the relatively erroneous ego-motion initialization.


\subsubsection{Incremental Approach in SE(2) with Scale-Initialized Odometry:} \label{subsubsec:bmvc}
A variant of the multibody monocular pose-graph based optimization pipeline defined in an SE(2) world. This approach optimizes for multiple objects in each frame in an incremental manner. There is a feedback of ``optimization results" back as the input to the optimizer in the next iteration in this approach. The parameterization of the pose-graph optimizer reduces quite considerably in this approach when compared with Sec.~\ref{subsubsec:baseline2} as we now function on a world governed by $3$ degrees of freedom as opposed to $6$. 

\subsection{Qualitative Results}
\label{subsec:qualres}

We obtain accurate localizations to vehicles in the camera's view using Sec.~\ref{subsec:vehloc}. The results have been illustrated as tight and accurate 3D bounding boxes obtained to the vehicles in Fig.~\ref{fig:pseudolidar_qualitative_results}. Fig.~\ref{fig:main_qualitative_results} illustrates the trajectories obtained after pose-graph optimization led to accurate bird's-eye view mappings of the ego car and the dynamic vehicles localized in the scene in a stationary world frame. Despite having to cope with a high error contributed to by the motion model predictor from Sec.~\ref{subsec:egomot}, we obtain close to ground truth \bev{} mapping post-optimization.  

\subsection{Quantitative Results}
\label{subsec:quantres}

Table.~\ref{table:main_comparison_odometry} and Table.~\ref{table:main_comparison_vehicle} presents the quantitative performance on a comparative footing for the ego car and the vehicles localized in the camera's scene respectively. Our batch-version with scale-ambiguous odometry initialization showcases a much superior performance compared with other batch-version, such as Namdev \etal~\cite{Namdev} and Nair \etal~\cite{iv}. Our batch-approach beats them for all but one vehicle shown in Table.~\ref{table:main_comparison_vehicle}. On the incremental version's front, we compare our \bev{} approach with the corresponding baseline defined in SE(3) as well as CubeSLAM~\cite{cubeslam}, whose errors are computed accordingly after running the codebase released by the respective authors on the particular sequence for which the input data and the tuned parameters(for that sequence) were made available. While the performance is comparable between the SE(3) as well as the \bev{} approach, we put ahead a much superior performance with respect to CubeSLAM~\cite{cubeslam}. This supports the statement that a \bev{} SLAM approach can potentially perform as well as its SE(3) counterpart.

\begin{table*}[!htbp]
\caption{Analysis over the contribution of each type of constraint to the final cost.}
\label{table:change name 3} \centering
\begin{center}
\begin{adjustbox}{max width=0.6\linewidth}
\begin{tabular}{|c|c||c|c|c|c|c|}

\hline
\multirow{2}{*}{Seq No.} & \multirow{2}{*}{Car ID} & \multicolumn{5}{c|}{Absolute Translation Error (RMS) in Global Frame (meters)} \\

\cline{3-7}
 &  & Without CC & Without CV & Without VV & Without CP & With all \\ \hline \hline

\multirow{2}{*}{2} & $1$ & $3.41$ & $1.96$ & $\mathbf{1.86}$ & $1.95$ & $2.09$ \\ \cline{2-7}

& Ego & $2.95$ & $2.25$ & $2.25$ & $\mathbf{2.15}$ & $2.25$ \\ \hline

\multirow{3}{*}{3} & $0$ & $2.35$ & $2.40$ & $2.69$ & $2.46$ & $\mathbf{2.37}$ \\ \cline{2-7}

& $1$ & $2.74$ & $2.05$ & $2.23$ & $2.12$ & $\mathbf{2.05}$ \\ \cline{2-7}

& Ego & $2.73$ & $1.78$ & $1.78$ & $1.96$ & $\mathbf{1.78}$ \\ \hline

\multirow{2}{*}{4} & $2$ & $4.52$ & $2.45$ & $\mathbf{2.26}$ & $2.58$ & $2.34$ \\ \cline{2-7}

& Ego & $6.82$ & $6.60$ & $6.60$ & $\mathbf{6.42}$ & $6.60$ \\ \hline

\multirow{2}{*}{5} & $31$ & $3.43$ & $2.01$ & $1.98$ & $1.98$ & $\mathbf{1.98}$ \\ \cline{2-7}

& Ego & $3.05$ & $1.58$ & $1.58$ & $\mathbf{1.57}$ & $1.58$ \\ \hline

\multirow{2}{*}{10} & $0$ & $15.72$ & $\mathbf{2.81}$ & $2.99$ & $2.98$ & $3.03$ \\ \cline{2-7}

& Ego & $15.43$ & $2.99$ & $3.52$ & $3.00$ & $\mathbf{2.99}$ \\ \hline

\multirow{4}{*}{18} & $1$ & $\mathbf{1.27}$ & $1.65$ & $1.30$ & $1.50$ & $1.60$ \\ \cline{2-7}

& $2$ & $2.90$ & $2.77$ & $2.84$ & $2.96$ & $\mathbf{2.76}$ \\ \cline{2-7}

& $3$ & $1.63$ & $1.82$ & $2.13$ & $1.74$ & $\mathbf{1.60}$ \\ \cline{2-7}

& Ego & $2.25$ & $2.21$ & $2.21$ & $2.24$ & $\mathbf{2.21}$ \\ \hline

\multirow{3}{*}{20} & $12$ & $12.35$ & $8.75$ & $9.33$ & $8.69$ & $\mathbf{8.61}$ \\ \cline{2-7}

& $122$ & $17.65$ & $10.32$ & $10.57$ & $\mathbf{10.09}$ & $10.12$ \\ \cline{2-7}

& Ego & $13.85$ & $8.86$ & $8.86$ & $8.88$ & $\mathbf{8.86}$ \\ \hline

\end{tabular}
\end{adjustbox}
\end{center}
\end{table*}


\begin{table*}[!htbp]
\caption{Performance of the optimiser as a function of weight given to the landmark based constraints relative to the same for ego motion [rows 1 - 3]. Performance of the optimiser with respect to the threshold set for the static feature landmarks on their depth from the camera [rows 4 - 8].}
\label{table:change name 4} \centering
\begin{center}
\begin{adjustbox}{max width=0.6\linewidth}
\begin{tabular}{|c|c||c|c|c||c|c|c|c|c|}

\hline
\multirow{3}{*}{Seq No.} & \multirow{3}{*}{Car ID} & \multicolumn{8}{c|}{Absolute Translation Error (RMS) in Global Frame (meters)} \\

\cline{3-10}
&  & \multicolumn{3}{c|}{Weight to CP} & \multicolumn{5}{c|}{Depth Threshold T (m)} \\ 

\cline{3-10}
&  & Low & Medium & High & $12$ & $15$ & $18$ & $20$ & $\infty$ \\ 
\hline \hline

\multirow{2}{*}{2} & $1$ & $2.14$ & $\mathbf{2.09}$ & $3.15$ & $2.27$ & $2.40$ & $2.14$ & $\mathbf{2.09}$ & $2.00$ \\ \cline{2-10}

& Ego & $\mathbf{2.16}$ & $2.25$ & $2.82$ & $2.36$ & $2.40$ & $2.32$ & $\mathbf{2.25}$ & $2.28$ \\ \hline

\multirow{3}{*}{3} & $0$ & $2.46$ & $2.37$ & $\mathbf{2.35}$ & $2.31$ & $2.37$ & $\mathbf{2.31}$ & $2.37$ & $2.40$ \\ \cline{2-10}

& $1$ & $2.12$ & $\mathbf{2.05}$ & $2.76$ & $2.05$ & $2.10$ & $2.14$ & $\mathbf{2.05}$ & $2.07$ \\ \cline{2-10}

& Ego & $1.96$ & $\mathbf{1.78}$ & $2.76$ & $1.80$ & $1.82$ & $1.80$ & $\mathbf{1.78}$ & $1.87$ \\ \hline

\multirow{2}{*}{4} & $2$ & $\mathbf{2.17}$ & $2.34$ & $4.67$ & $2.54$ & $2.34$ & $2.36$ & $2.34$ & $\mathbf{2.26}$ \\ \cline{2-10}

& Ego & $\mathbf{6.42}$ & $6.60$ & $5.72$ & $6.59$ & $\mathbf{6.42}$ & $6.58$ & $6.60$ & $6.45$ \\ \hline

\multirow{2}{*}{5} & $31$ & $1.98$ & $\mathbf{1.98}$ & $2.05$ & $1.98$ & $1.98$ & $1.98$ & $1.98$ & $\mathbf{1.90}$ \\ \cline{2-10}

& Ego & $1.60$ & $\mathbf{1.58}$ & $1.67$ & $1.58$ & $\mathbf{1.57}$ & $1.58$ & $1.58$ & $1.60$ \\ \hline

\multirow{2}{*}{10} & $0$ & $\mathbf{2.99}$ & $3.03$ & $3.30$ & $3.09$ & $3.11$ & $3.03$ & $\mathbf{3.03}$ & $3.18$ \\ \cline{2-10}

& Ego & $\mathbf{2.96}$ & $2.99$ & $3.19$ & $3.01$ & $3.00$ & $2.99$ & $\mathbf{2.99}$ & $3.07$ \\ \hline

\multirow{4}{*}{18} & $1$ & $\mathbf{1.50}$ & $1.60$ & $1.28$ & $1.59$ & $\mathbf{1.59}$ & $1.65$ & $1.60$ & $1.68$ \\ \cline{2-10}

& $2$ & $2.96$ & $2.76$ & $\mathbf{2.75}$ & $2.91$ & $2.87$ & $2.83$ & $\mathbf{2.76}$ & $2.85$ \\ \cline{2-10}

& $3$ & $1.74$ & $\mathbf{1.60}$ & $1.80$ & $1.74$ & $1.66$ & $1.62$ & $\mathbf{1.60}$ & $1.66$ \\ \cline{2-10}

& Ego & $2.24$ & $2.21$ & $\mathbf{2.15}$ & $2.45$ & $2.26$ & $2.25$ & $\mathbf{2.21}$ & $2.25$ \\ \hline

\multirow{3}{*}{20} & $12$ & $8.70$ & $\mathbf{8.61}$ & $9.38$ & $8.72$ & $8.74$ & $8.68$ & $\mathbf{8.61}$ & $8.64$  \\ \cline{2-10}

& $122$ & $\mathbf{10.09}$ & $10.12$ & $10.36$ & $10.12$ & $\mathbf{10.10}$ & $10.18$ & $10.12$ & $10.17$ \\ \cline{2-10}

& Ego & $8.90$ & $\mathbf{8.86}$ & $9.63$ & $8.88$ & $8.90$ & $\mathbf{8.84}$ & $8.86$ & $8.85$ \\ \hline

\end{tabular}
\end{adjustbox}
\end{center}
\end{table*}

\begin{table*}[!htbp]
\caption{Impact of lane-based constraints on batch-based approach from Sec.~\ref{subsubsec:baseline2}.}
\label{table:ablation_lane_constraint} \centering
\begin{center}
\begin{adjustbox}{max width=0.8\linewidth}
\begin{tabular}{|c|c|c|c|c|c|c|c|c|c|c|}

\hline
& \multicolumn{10}{c|}{Absolute Translation Error (RMS) in Global Frame (meters)} \\ \hline

Seq No. & \multicolumn{3}{c|}{3} & \multicolumn{2}{c|}{4} & \multicolumn{4}{c|}{18} & \multirow{3}{*}{Avg Error} \\ \cline{1-10}

Car ID & $0$ & $1$ & Ego-car & $2$ & Ego-car & $1$ & $2$ & $3$ & Ego-car &\\ \cline{1-10}

Frame length & $41$ & $92$ & $123$ & $149$ & $149$ & $62$ & $83$ & $141$ & $141$ & \\ \hline

Before Lane-Constraints &	$2.91$ & $2.61$ & $2.26$ & $2.15$ & $4.82$ & $1.32$ & $3.22$ & $\mathbf{1.19}$	& $2.53$ & $2.56$ \\ \hline

After Lane-Constraints & $\mathbf{2.20}$	& $\mathbf{2.24}$ & $\mathbf{1.96}$ & $\mathbf{1.77}$ &	$\mathbf{1.89}$ & $\mathbf{1.21}$ & $\mathbf{2.86}$ & $1.23$ & $\mathbf{2.36}$ & $\mathbf{1.97}$ \\ \hline

\end{tabular}
\end{adjustbox}
\end{center}
\end{table*}

\subsection{Ablation Studies on Real-Time Approaches}
\label{subsec:analysis-rt}

\subsubsection{Contribution by Individual Constraints}
\label{subsubsec:edgecontri}

We analyze each constraint's contribution as summarized in Sec.~\ref{subsubsec:constr} by computing the final error after allotting zero weight to individual constraints, effectively removing its influence on the optimization. The observations are presented in Table.~\ref{table:change name 3}. Since the $CC$ constraints are given high weight, as explained in Sec.~\ref{subsubsec:constr}, the removal of this constraint results in the deterioration of performance for ego-motion. It can also be seen that, through the $CP$ constraints, the stationary points help enhance ego-motion in most cases. The $CV$ edge that primarily utilizes the pseudolidar~\cite{pseudolidar} based localization ensures that the relation between the ego-motion and all the vehicles in its scene remains synchronized. 

\subsubsection{Weight Allotted to Landmark Based Constraints}
\label{subsubsec:weightP}

While it has been established from Sec.~\ref{subsubsec:edgecontri} that static landmarks help improve the absolute translation error of the trajectory, we analyze as to how much emphasis must be given to the CP constraint in terms of the weight. We experiment with various levels of weights fed to the CP constraint in relation with that of the CC edge in the formulation. Table.~\ref{table:change name 4} summarizes our observations. While medium weight, which is equal to that of CC constraints, beats other modes by a huge margin in a few instances, it competes closely in all the other instances. On the whole, the performance put forth with medium weight to CP constraints is superior to the other modes.

\subsubsection{Threshold for Landmarks}
\label{subsubsec:thresholdP}

Since point correspondences and the depth estimations to the same may be more reliable for features closer to the camera, we place a threshold along Z-axis of the camera to shortlist landmarks to be considered in CP constraints as mentioned in Sec.~\ref{subsubsec:constr}. Our experiments with various thresholds have been reported in Table.~\ref{table:change name 4}. We find that a threshold $T = 20m$ contributes optimally to the pose-graph optimization step.

\subsubsection{Impact of Lane Constraints}
\label{subsubsec:lane_constraint_impact}

We show ablation studies on \emph{lane-based constraining} of trajectories in our batch-based pose-graph formulation from Sec. \ref{subsubsec:baseline2}. These are performed on unscaled-ORB initializations. We show that lane-constraints contribute by with substantial improvement in ATE for almost all vehicles which are experimented with, when compared with the corresponding ATE before applying lane-based constraints. We summarize our observations in Table.~\ref{table:ablation_lane_constraint}.

\subsubsection{Runtime Analysis}
\label{subsubsec:runtime}

\begin{figure}[htbp]
    \centering
        \includegraphics[width=0.9\linewidth]{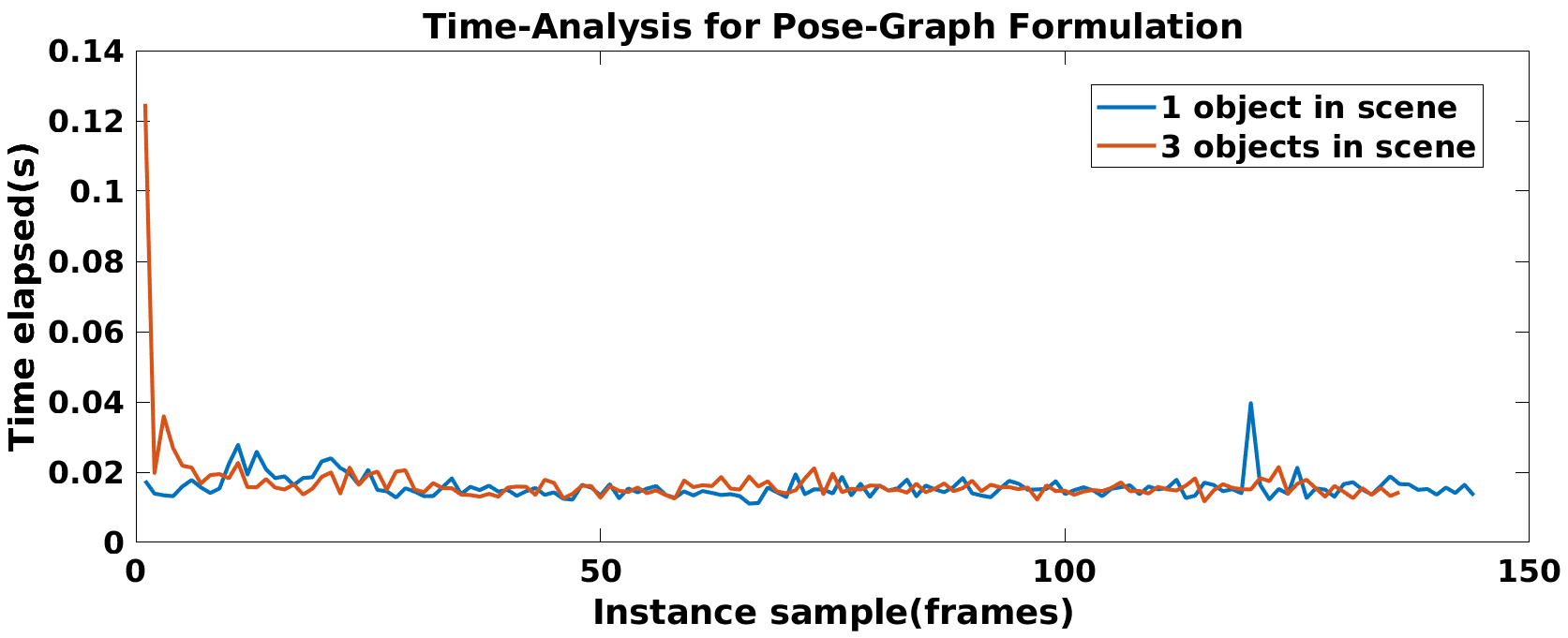}
    \caption{Plot illustrating how number of objects in scene do not affect the time-elapsed in our optimization formulation from Sec. \ref{subsec:posegraph}}
    \label{fig:timePlot}
\end{figure}

The incremental optimizer in Sec.\ref{subsubsec:bmvc} takes $0.016$s to solve the pose-graph optimization problem for a $414$ frame long sequence as compared to $1.9$s for batch-based approach in Sec.\ref{subsubsec:baseline2}. Fig. \ref{fig:timePlot} shows how a single and multi-object scenario fare in terms of runtime for each incoming instance. Pose-graph optimizations (see \ref{subsec:posegraph}) are performed on a quadcore Intel i7-5500U CPU with $2.40GHz$ processor. The frontend involves gathering predictions from multiple neural networks~\cite{monodepth2,pseudolidar,frustum_pointnet} and runs at around $33$ Hz frequency.

\section{CONCLUSION}
\label{sec:conclusion}

Multibody SLAM in a moving monocular setup is a difficult problem to solve given its \emph{ill-posedness}.
In this paper, we operate in an orthographic (\bev{}) space to overcome the challenges posed by dynamic scenes to the conventional monocular SLAM systems. 
Moreover, \birdslam{} operates in real-time in \bev{} space performing better than current real-time state-of-the-art multibody SLAM systems operating in 6 DoF setup. It also performs at par with current offline multibody SLAM systems operating under strictly more resources (time, computation, features). To the best of our knowledge, \birdslam{} is the one of the first such system to demonstrate a solution to the multibody monocular SLAM problem in orthographic space. An interesting future direction could be to consider cases in which the single-view metrology cues do not hold, such as on extremely graded/steep roads.

\bibliographystyle{apalike}
{\small
\bibliography{references}}

\end{document}